\documentclass[10pt,twocolumn,letterpaper]{article}

\usepackage{cvpr}
\usepackage{times}
\usepackage{epsfig}
\usepackage{graphicx}
\usepackage{amsmath}
\usepackage{amssymb}
\usepackage{multirow}
\usepackage{algorithm}
\usepackage{algpseudocode}
\usepackage{rotating}
\usepackage{cite}
\newcommand*\rot{\rotatebox{90}}
\usepackage{colortbl}
\usepackage{soul}
\definecolor{Gray}{gray}{0.9}
\newcolumntype{g}{>{\columncolor{Gray}} c}

\usepackage[breaklinks=true,letterpaper=true,colorlinks,bookmarks=false,pagebackref=true]{hyperref}

\cvprfinalcopy 

\def\cvprPaperID{383} 

\ifcvprfinal\fi
\begin{document}

\title{Large-Scale Visual Active Learning with Deep Probabilistic Ensembles}

\author{Kashyap Chitta\\
Carnegie Mellon University\\
{\tt\small kchitta@andrew.cmu.edu}
\and
Jose M. Alvarez\\
NVIDIA\\
{\tt\small josea@nvidia.com}
\and
Adam Lesnikowski\\
NVIDIA\\
{\tt\small alesnikowski@nvidia.com}
}

\maketitle

\begin{abstract}
Annotating the right data for training deep neural networks is an important challenge. Active learning using uncertainty estimates from Bayesian Neural Networks (BNNs) could provide an effective solution to this. Despite being theoretically principled, BNNs require approximations to be applied to large-scale problems, where both performance and uncertainty estimation are crucial. In this paper, we introduce Deep Probabilistic Ensembles (DPEs), a scalable technique that uses a regularized ensemble to approximate a deep BNN. We conduct a series of large-scale visual active learning experiments to evaluate DPEs on classification with the CIFAR-10, CIFAR-100 and ImageNet datasets, and semantic segmentation with the BDD100k dataset. Our models require significantly less training data to achieve competitive performances, and steadily improve upon strong active learning baselines as the annotation budget is increased.
\end{abstract}

\section{Introduction}
Collecting the right data for supervised deep learning is an important and challenging task that can improve performance on most modern computer vision problems. Active learning aims to select, from a large unlabeled dataset, the smallest possible training set to solve a specific task \cite{cohn1994active}. To this end, the \textit{uncertainty} of a model is used as a means of quantifying what the model does not know, to select data to be annotated. In deterministic neural networks, uncertainty is typically measured as the confidence in the output of the last softmax-normalized layer. However, these estimates do not always provide reliable information to select appropriate training data, as neural networks tend to be overconfident \cite{szegedy2014intriguing,guo2017calibration}. On the other hand, Bayesian methods provide a principled approach to estimate the uncertainty of a model. Though they have recently gained momentum, they have not found widespread use in practice \cite{kendall2015bayesian,smith2018understanding,gal2018sufficient}.

The formulation of a Bayesian Neural Network (BNN) involves placing a prior distribution over all the parameters of the network, and obtaining the posterior given the observed data \cite{neal1995bayesian}. The spread in the {distribution of predictions} provided by a trained BNN helps to capture the model's uncertainty. However, training a BNN involves marginalization over all possible assignments of weights, which is intractable for deep BNNs without approximations \cite{graves2011practical,blundell2015weight,gal2015bayesian}. Existing approximation algorithms limit their applicability, since they do not specifically address the fact that deep BNNs on large datasets are more difficult to optimize than deterministic networks, and require extensive parameter tuning to provide good performance and uncertainty estimates \cite{osband2016risk}. Furthermore, estimating uncertainty in BNNs requires drawing a large number of samples at test time, which can be extremely computationally demanding \cite{gal2015bayesian}.

In practice, a common approach to estimate uncertainty is based on ensembles \cite{lakshminarayanan2017simple,beluch2018power,geifman2018boosting}. Different models in an ensemble of networks are treated as if they were samples drawn directly from a BNN posterior. Ensembles are easy to optimize and fast to execute. However, they do not approximate uncertainty in the same manner as a BNN. For example, the parameters in the first kernel of the first layer of a convolutional neural network may serve a completely different purpose in different members of the ensemble. Therefore, the variance of the values of these parameters after training cannot be compared to the variance that would have been obtained in the first kernel of the first layer of a trained BNN with the same architecture.

In this paper, we propose Deep Probabilistic Ensembles (DPEs), a novel approach to regularize ensembles based on BNNs. Specifically, we use variational inference \cite{blei2016variational}, a popular technique for training BNNs, to derive a generic regularization approach for ensembles. Our formulation focuses on a specific form of this regularization designed for active learning, leading to promising results on large-scale visual classification benchmarks with up to millions of samples. Our technique has \textit{no computational overhead} compared to ensembles at inference time, and a negligible overhead during training. When applied to CIFAR-10, CIFAR-100 and ImageNet, DPEs consistently outperform strong baselines. To further show the applicability of our method, we propose a framework to actively annotate data for semantic segmentation on the challenging BDD100k dataset, which on certain classes that are underrepresented in the training distribution yields IoU improvements of up to 20\%.

Our contributions therefore are (i) we propose KL regularization for training DPEs, which combine the advantages of ensembles and BNNs; (ii) we apply DPEs to active image classification on large-scale datasets; and (iii) we propose a framework for active semantic segmentation using DPEs. Our experiments on both classification and segmentation demonstrate the benefits and scalability of our method compared to existing methods. DPEs are parallelizable, easy to implement, yield high performance, and provide good uncertainty estimates when used for active learning.

\section{Related Work}

\noindent
\textbf{Regularizing Ensembles.} Using ensembles of neural networks to improve performance has a long history \cite{hansen1990neural,dietterich2000ensemble}. Attempts to regularize ensembles have typically focused on promoting diversity among the ensemble members, through penalties such as negative correlation \cite{liu1999ensemble}. While these methods focus on improving the accuracy, our approach instead focuses on the ensemble's uncertainty estimates.

\noindent
\textbf{Training BNNs.} There are four popular approximations for training BNNs. The first one is \textit{Markov Chain Monte Carlo} (MCMC). MCMC approximates the posterior through Bayes rule, by sampling to compute an integral \cite{robert2005montecarlo}. This method is accurate but it is extremely sample inefficient, and has not been applied to deep BNNs. The second one is \textit{Variational Inference} which is an iterative optimization approach to update a chosen variational distribution on each network parameter using the observed data \cite{blei2016variational}. Training is fast for reasonable network sizes, but performance is not always ideal, especially for deeper networks, due to the nature of the optimization objective.

The third approximation is \textit{Monte Carlo Dropout} (MC Dropout) \cite{gal2017deep} (and its variants \cite{teye2018bayesian,atanov2018uncertainty}). These methods regularize the network during training with a technique like dropout \cite{srivastava2014dropout}, and draw Monte Carlo samples \textit{during test time} with different dropout masks as approximate samples from a BNN posterior. It can be seen as an approximation to variational inference when using Bernoulli priors \cite{gal2015bayesian}. Due to its simplicity, several approaches have been proposed using MC dropout for uncertainty estimation \cite{kendall2015bayesian,kendall2017uncertainties,gal2016Uncertainty,gal2018sufficient}. The last approximation is  \textit{Bayes by Backprop} \cite{blundell2015weight} which involves maintaining two parameters on each weight, a mean and standard deviation, with their gradients calculated by backpropagation. Empirically, the uncertainty estimates obtained with this approach and MC Dropout perform similarly for active learning \cite{siddhant2018deep}. 

Unfortunately, the uncertainty estimates obtained with these BNN approximation techniques do not perform as well in practice as those obtained from ensembles on large-scale visual tasks \cite{lakshminarayanan2017simple,beluch2018power}. In addition, MC dropout requires the dropout rate to be very well-tuned to produce reliable uncertainty estimates \cite{osband2016risk}. Given the success of BNN-based techniques for active learning in the small-scale setting, our approach aims to introduce BNN-like uncertainty propagation in ensembles, to improve their active learning performance. 

\noindent
\textbf{(Deep) Active Learning.} A comprehensive review of classical approaches to active learning can be found at \cite{settles2010active}. Most of these approaches rely on a confusion metric based on the model's predictions, like information theoretical methods that maximize expected information gain \cite{mackay1992information}, committee based methods that maximize disagreement between committee members \cite{dagan1995committee,freund1997selective,mccallum1998employing,culotta2005reducing}, entropy maximization methods \cite{lewis1994sequential}, and margin based methods \cite{schohn2000less,campbell2000query,tong2002support,joshi2009multiclass}. Data samples for which the current model is uncertain, or {most confused}, are queried for labeling usually one at a time and, in general, the model is re-trained for each of these queries. In the era of deep neural networks, querying single samples and retraining the model for each one is not feasible. Adding a single sample into the training process is likely to have no statistically significant impact on the performance of a deep neural network and the retraining becomes computationally intractable.

Scalable alternatives to classical active learning approaches have been explored for the iterative batch query setting. Pool-based approaches filter unlabeled data using heuristics based on a confusion metric, and then choose what to label based on diversity within this pool \cite{demir2011batchmode,wang2017cost,zhang2017active}. More recent approaches explore active learning without breaking the task down using confusion metrics, through a lower bound on the prediction error known as the core-set loss \cite{sener2018active} or learning the interestingness of samples in a data-driven manner through reinforcement learning \cite{woodward2017active,fang2017learning,pang2018metalearning}. Results in these directions are promising. However, state-of-the-art active learning algorithms are based on uncertainty estimates from network ensembles \cite{beluch2018power}. Our experiments demonstrate the gains obtained over ensembles when they are trained with the proposed regularization scheme.

\section{Deep Probabilistic Ensembles}

\subsection{Variational Inference}

For inference in a BNN, we can consider the weights $w$ to be latent variables drawn from our prior distribution, $p(w)$. These weights relate to the observed dataset $x$ through the likelihood, $p(x|w)$. We aim to compute the posterior,
\begin{equation}
p(w|x) = \frac{p(w)p(x|w)}{p(x)} = \frac{p(w)p(x|w)}{\int p(w) p(x|w) dw}.
\label{e:bayes}
\end{equation}

It is intractable to analytically integrate over all assignments of weights, so variational inference restricts the problem to a family of distributions $D$ over the latent variables. Within this family, a solution is to optimize for the member $q^*(w)$ that is closest to the true posterior in terms of KL divergence,
\begin{align}
q^*(w) & = \underset{q(w) \in D}{\arg\min} \ KL(q(w)||p(w|x)) \\
& = \underset{q(w) \in D}{\arg\min} \ \mathbb{E} \left[ \log \frac{q(w)}{p(w|x)} \right].
\end{align}

To make the computation tractable the objective is modified by subtracting a constant independent of the weights $w$. This new objective to be minimized is the negative of the Evidence Lower Bound (ELBO) \cite{blei2016variational},
\begin{align}
-ELBO & = \mathbb{E}\left[\log \frac{q(w)}{p(w|x)}\right] - \log p(x)\\
 & = \mathbb{E}[\log q(w)] - \mathbb{E}[\log p(w|x)] - \log p(x).\nonumber
\end{align}

Using the posterior term from Eq. \ref{e:bayes}, this can be simplified to
\begin{align}
-ELBO & = \mathbb{E}[\log q(w)] - \mathbb{E}[\log p(w)p(x|w)] \nonumber\\
 & = \mathbb{E}\left[\log \frac{q(w)}{p(w)}\right] - \mathbb{E}[\log p(x|w)] \nonumber\\
 & = KL(q(w)||p(w)) - \mathbb{E}[\log p(x|w)], 
 \label{e:elbo}
\end{align}
\noindent where the first term is the KL divergence between $q(w)$ and a chosen prior distribution $p(w)$; and the second term is the expected negative log likelihood (NLL) of the data $x$ based on the current parameters $w$. The optimization difficulty in variational inference arises partly due to this expectation of the NLL. In deterministic networks, fragile co-adaptations exist between different parameters, which can be crucial to their performance \cite{yosinski2014transferable}. Features typically interact with each other in a complex way, such that the optimal setting for certain parameters is highly dependent on specific configurations of the other parameters in the network. Co-adaptation makes training easier, but can reduce the ability of deterministic networks to generalize \cite{hinton2012improving}. Popular regularization techniques to improve generalization, such as Dropout, can be seen as a trade-off between co-adaptation and training difficulty \cite{srivastava2014dropout}. An ensemble of deterministic networks exploits co-adaptation, as each network in the ensemble is optimized independently, making them easy to optimize.

For BNNs, the nature of the objective, an expectation over all possible assignments of $q(w)$, prevents the BNN from exploiting co-adaptations during training, since we seek to minimize the NLL for \textit{any generic deterministic network} sampled from the BNN. While in theory this should lead to great generalization, in practice, it becomes very difficult to tune a BNN to produce competitive results. In this paper, we propose a form of regularization to use the optimization simplicity of ensembles for training BNNs. The main properties of our approach compared to ensembles and BNNs are summarized in Table \ref{t:coadapt}.

\begin{table}[!t]
	\begin{center}
    \caption{Properties of the proposed approach (DPE) compared to BNNs and ensembles. DPEs are simple to optimize like ensembles, and have a training objective similar to BNNs. As a result, we obtain high performance and good uncertainty estimates.}
    \label{t:coadapt}
		\begin{tabular}{c|c|c|c}
			\hline
			{Model} & Co-adaptation & Training & Generalization\\
            \hline
            Ensemble & High & Easy & Low\\
            DPE & Medium & Easy & High\\
            BNN & Low & Hard & High\\
            \hline
        \end{tabular}
	\end{center}	
	\vspace{-0.5cm}
\end{table}

\subsection{KL Regularization}
The standard approach to training neural networks involves regularizing each individual parameter with $L_1$ or $L_2$ penalty terms. We instead apply the KL divergence term in Eq. \ref{e:elbo} as a regularization penalty $\Omega$, to the \textit{set of values} that a given parameter takes over \textit{all members in an ensemble}. The choice of prior distribution depends on the task-- for example, a log-uniform prior induces sparsity and can be used for pruning \cite{achterhold2018variational}. For our task, we choose the family of Gaussian functions for $p(w)$ and $q(w)$, which allows the KL divergence to be analytically computed by assuming mutual independence between the network parameters and factoring the term into individual Gaussians \cite{blei2016variational}. The KL divergence between two Gaussians with means $\mu_q$ and $\mu_p$, standard deviations $\sigma_q$ and $\sigma_p$ is given by
\begin{equation}
KL(q||p) = \frac{1}{2}\left(\log \frac{\sigma_q^2}{\sigma_p^2} + \frac{\sigma_p^2 + (\mu_q - \mu_p)^2}{\sigma_q^2} - 1\right).
\label{e:kl}
\end{equation}

\noindent
However, the distribution we obtain from the {set of values} that a given parameter takes over {all members in an ensemble} is a mixture of Dirac delta functions, and not a Gaussian. We therefore use maximum likelihood estimates for the mean and standard deviation, $\mu_q$ and $\sigma_q$, from this distribution to approximate $q(w)$.

In our experiments, we choose the network initialization technique proposed by He et al. \cite{he2015delving} as a prior. Specifically, for batch normalization parameters, we fix $\sigma_p^2=0.01$, and set $\mu_p=1$ for the weights and $\mu_p=0$ for the biases. For the weights in convolutional layers with the ReLU activation (with $n_i$ input channels, $n_o$ output channels, and kernel dimensions $w \times h$) we set $\mu_p=0$ and $\sigma_p^2=\frac{2}{n_{o}wh}$. Linear layers can be considered a special case of convolutions, where the kernel has the same dimensions as the input activation. The KL regularization of a layer $\Omega^l$ is obtained by removing the terms independent of $q$ and substituting for $\mu_p$ and $\sigma_p$ in Eq. \ref{e:kl},
\begin{equation}
\Omega^l = \sum_{i=1}^{n_{i} n_{o} w h} \left(\log {\sigma_i^2} + \frac{2}{n_{o} w h\sigma_i^2} + \frac{\mu_i^2}{\sigma_i^2}\right),
\label{e:omega}
\end{equation}
\noindent where $\mu_i$ and $\sigma_i$ are the mean and standard deviation of the set of values taken by a parameter across different ensemble members. In Eq. \ref{e:omega}, the first term prevents extremely large variances compared to the prior, so the ensemble members do not diverge completely from each other. The second term heavily penalizes variances less than the prior, promoting diversity between members. The third term keeps the mean of the weights close to that of the prior, especially when their variance is also low.

\noindent
\textbf{Objective function.} We can now rewrite the minimization objective in Eq. \ref{e:elbo} for an ensemble as:
\begin{equation}
\mathbf{\Theta}^* = \underset{\mathbf{\Theta}}{\arg\min} \sum_{i=1}^{N} \sum_{e=1}^{E} \mathcal{H}(y^i,\mathcal{M}_e(\textbf{x}^i,\Theta_e)) + \beta \Omega( \mathbf{\Theta}),
\end{equation}
where $\{(\textbf{x}^i,y^i)\}_{i=1}^{N}$ is the training data, $E$ is the number of models in the ensemble, and ${\Theta}_e$ refers to the parameters of the model $\mathcal{M}_e$. $\mathcal{H}$ is the cross-entropy loss for classification and $\Omega$ is our KL regularization penalty over all the parameters of the ensemble $\mathbf{\Theta}$. We obtain $\Omega$ by aggregating the penalty from Eq. \ref{e:omega} over all the layers in a network, and use a scaling term $\beta$ to balance the regularizer with the likelihood loss. By summing the loss over each independent model, we are approximating the expectation of the ELBO's NLL term in Eq. \ref{e:elbo} over only the current ensemble configuration, a subset of all possible assignments of $q(w)$. This is the main distinction between our approach and traditional variational inference.

\section{Uncertainty Estimation for Active Learning}
In this section, we describe how DPEs can be used for active learning. Our pipeline consists of four main components:
\begin{itemize}
	\item A large \textbf{unlabeled set} consisting of $N_u$ samples, $U=\{\textbf{x}_u^j\}_{j=1}^{N_u}$, where each $\textbf{x}^j \in X$ is an input data point. 
    \item A \textbf{labeled set}, consisting of $N_l$ labeled pairs, $L=\{(\textbf{x}_l^j,y_l^j)\}_{j=1}^{N_l}$, , where each $\textbf{x}^j \in X$ is a data point and each $y^j \in Y$ is its corresponding label. For a $K$-way classification problem, the label space $Y$ would be $\{1, ..., K\}$.
    \item An \textbf{annotator}, $\mathcal{A}:X\rightarrow Y$ that can be queried to map data points to labels
    \item Our \textbf{DPE} $\mathcal{M}$, a set of models $\{\mathcal{M}_e:X\rightarrow Y\}_{e=1}^{E}$, each trained to estimate the label of a given data point.
\end{itemize}

These four components interact with each other in an iterative process. In our setup, commonly referred to as \textit{batch mode active learning} \cite{guo2008discriminative}, every iteration involves 2 stages, which are summarized in Algorithm \ref{a:flow}. In the first stage, the current labeled set $L^{(i)}$ is used to train the ensemble $\mathcal{M}^{(i)}$ to convergence. In the second stage, the trained ensemble is applied to the existing unlabeled pool ${U}^{(i)}$, to select a subset of $b$ samples from this pool to send to the annotator $\mathcal{A}$ for labeling. This growth parameter $b$ may be fixed or vary over iterations. The subset is moved from $U^{(i)}$ to $L^{(i+1)}$ before the next iteration. Training proceeds in this manner until an annotation budget of $B$ total label queries is met.

\begin{algorithm}[!t]
\caption{Active learning with DPEs.}
\label{a:flow}
\small{
\begin{algorithmic}
    \State $i \leftarrow 0$
    \State Randomly sample $b$ points $\{\textbf{x}^j\}_{j=1}^{b}$ from unlabeled set $U^{(0)}$
    \State Annotate this data, $\{y^j = \mathcal{A}(\textbf{x}^j) \}_{j=1}^{b}$
    \State Move $\{(\textbf{x}^j,y^j)\}_{j=1}^{b}$ to initial labeled set $L^{(0)}$
    \State total\_added\_samples $\leftarrow b$
    \While{total\_added\_samples $\leq B$} 
       \While{model\_not\_converged} \Comment{\textbf{stage (i)}}
		\State Forward pass a mini-batch from $L^{(i)}$, $\hat{y} = \mathcal{M}^{(i)}(x)$
        \State Compute gradients $\delta$ of loss $l(y,\hat{y}) + \beta \Omega$
        \State Update DPE, $\mathcal{M}^{(i)} \leftarrow \mathcal{M}^{(i)} + \delta$
    \EndWhile
    \For{num\_iteration\_samples = 1 to $b$} \Comment{\textbf{stage (ii)}}
    \For{\textbf{x} $\in U^{(i)}$}
    \State $\alpha(\textbf{x}) \leftarrow H_{ens}(\textbf{x})$
    \EndFor
    \State $choice$ = $\arg \max_\textbf{x} \alpha(\textbf{x})$
    \State $y^{choice} = \mathcal{A}(\textbf{x}^{choice})$
	\State Move $(\textbf{x}^{choice},y^{choice})$ from $U^{(i)}$ to $L^{(i+1)}$
    \State total\_added\_samples += 1
    \EndFor
    \State Update $b$
    \State $i$ += 1
    \EndWhile
\end{algorithmic}
}
\end{algorithm}

Central to the second stage is the acquisition function $\alpha$ that is used for ranking all the available unlabeled data. For this, we employ the predictive entropy of the ensemble, given by
\begin{equation}
H_{ens} = -\left(\sum_{e \in E}\textbf{p}^{(e)}\right)^T\log \left(\sum_{e \in E}\textbf{p}^{(e)}\right),
\label{e:h_ens}
\end{equation}
\noindent where, $\textbf{p}$ refers to the normalized model prediction vector. Since the entropy for each query sample $\textbf{x} \in U^{(i)}$ is independent of the others, the entire batch of $b$ samples can be added to $L_{i+1}$ in parallel. This allows for increased efficiency, but we do not account for the fact that within the batch, adding certain samples has an impact on the uncertainty associated with the other samples.

In our experiments, to circumvent the need for annotators in the experimental loop, we use datasets for which all samples are annotated in advance, but hide the annotation away from the model. The dataset is initially treated as our unlabeled pool $U$, and labels are revealed to $\mathcal{M}$ as the experiment proceeds.

\section{Active Image Classification}
In this section, we evaluate the effectiveness of our approach on active learning for image classification across various levels of task complexity with respect to number of classes, image resolution  and quantity of data.

\noindent
\textbf{Datasets.} We experiment with three widely used benchmarks for image classification: the CIFAR-10 and CIFAR-100 datasets \cite{krizhevsky2009learning}, as well as the ImageNet 2012 classification dataset \cite{deng2009imagenet}. CIFAR datasets involve object classification tasks over natural images: CIFAR-10 is coarse-grained over 10 classes, and CIFAR-100 is fine-grained over 100 classes. For both tasks, there are 50k training images and 10k validation images of resolution $32 \times 32$. ImageNet consists of 1000 object classes, with annotation available for 1.28 million training images and 50k validation images of varying sizes and aspect ratios. All three datasets are balanced in terms of the number of training samples per class.

\noindent
\textbf{Network Architecture.} We  use  ResNet~\cite{he2016deep} backbones to  implement  the  individual  members  of  the  DPE. With ImageNet, we use a ResNet-18 with the standard kernel sizes and counts. For CIFAR, we use a variant of ResNet-18 as proposed in \cite{he2016identity}.

\begin{figure}[!t]
\centering
\includegraphics[width=0.95\columnwidth]{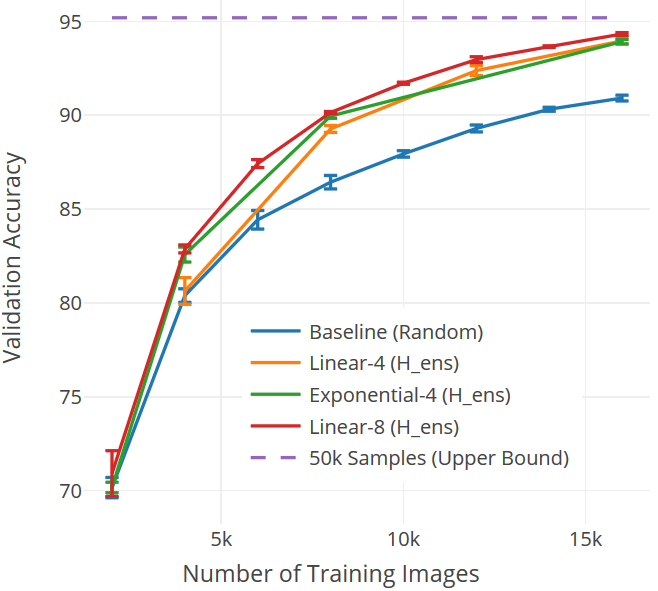}
\caption{\textbf{CIFAR-10:} Effect of varying dataset growth parameter $b$ on validation accuracy using the $H_{ens}$ acquisition function. Dashed line is the upper bound as the performance of the model trained with all the data. As shown, regardless of the value of $b$, our approach achieves competitive performance to this upper bound despite using only 32\% of the training data.}
\label{f:c10}
\vspace{-0.5cm}
\end{figure}

\noindent
\textbf{Implementation Details.} We retrain the ensemble from scratch for every iteration of active learning, to avoid remaining in a potentially sub-optimal local minimum from the earlier model trained with a smaller quantity of data. We do mean-std pre-processing, and augment the labeled dataset on-line with random crops and horizontal flips. Optimization is done using Stochastic Gradient Descent with a learning rate of 0.1, batch size of 32 and momentum of 0.9. The KL regularization penalty $\beta$ is set to $10^{-5}$. We use a patience parameter (set to 25) for counting the number of epochs with no improvement in validation accuracy, in which case the learning rate is dropped by a factor of 0.1. We end training when dropping the learning rate also gives no improvement in the validation accuracy after a number of epochs equal to twice the patience parameter. In each iteration, if the early stopping criterion is not met, we train for a maximum of 400 epochs. All our active classification experiments use the same hyper-parameters, without specific tuning.

In our experiments, we use ensembles of $E=8$ members. As long as the number of models is larger than 3, $\mu_i$ from Eq. \ref{e:omega} is nearly zero for most parameters, and training is stable. However, when using 2 models, there are several locations in the network for which $\mu_i > \sigma_i$. This causes exploding gradients due to the third term in Eq. \ref{e:omega}. In addition, we obtained no benefits when using a larger number of members than 8 (e.g., 16 or more).


\begin{table}[!t]
	\begin{center}
    \caption{Acquisition functions used in our second experiment.}
    \label{t:acquisition}
		\begin{tabular}{c|c}
		    \hline
		    \textbf{Function}&\textbf{Formula}\\
			\hline
			$H_{ens}$ & $-(\sum_{e \in E}\textbf{p}^{(e)})^T\log (\sum_{e \in E}\textbf{p}^{(e)})$ \\
            \hline
            $H_{cat}$ & $\sum_{e \in E}(-(\textbf{p}^{T(e)})\log (\textbf{p}^{(e)}))$ \\
            \hline
            $MI$ & $H_{ens}-H_{cat}$\\
            \hline
            $V$ &  $\sum_{k \in K} \underset{e \in E}{\text{Var}}(\textbf{p}_k^{(e)})$\\
            \hline
            $VR$ & $1 - \frac{f_m}{E}$\\
            & where $M = \underset{e \in E}{\text{Mode}}(\underset{k \in K}{\arg \max \ } \textbf{p}_k^{(e)})$\\
            & and $f_m = \sum_{e \in E} (\underset{k \in K}{\arg \max \ } \textbf{p}_k^{(e)} = M)$\\
            \hline
        \end{tabular}
	\end{center}	
	\vspace{-0.4cm}
\end{table}

\begin{table*}[!t]
	\begin{center}
    \caption{\textbf{Acquisition function on CIFAR:} Validation accuracy (in \%) of DPEs with different acquisition functions as defined in Table \ref{t:acquisition} for different labeling budgets. The initial 2k is randomly sampled. As shown, the final accucary is similar regardless the acquisition function.}
    \label{t:CIFAR-10}
		\begin{tabular}{l|c|c|c||c|c|c}
			\hline
			\small \textbf{Dataset}&\multicolumn{3}{c||}{\small \textbf{CIFAR-10}}&\multicolumn{3}{c}{\small \textbf{CIFAR-100}}\\\hline
			\small \textbf{Acquisition} & \small \textbf{4k (8\%)} & \small \textbf{8k (16\%)} & \small \textbf{16k (32\%)}& \small \textbf{4k (8\%)} & \small \textbf{8k (16\%)} & \small \textbf{16k (32\%)}\\
            \hline
			Random ($R$) & 80.60 & 86.80 & 91.08 & 39.57 & 54.92 & 66.65 \\
            Categories First Entropy ($H_{cat}$) & \textbf{82.96} & 89.92 & 94.10 & 38.47 & 55.63 & 69.16 \\
            Mutual Information ($MI$) & 82.70 & 90.00 & 93.97 & 39.96 & 55.50 & 69.01 \\
            Variance ($V$) & 82.19 & 90.05 & 93.92 & \textbf{41.13} & 56.62 & 69.17 \\
            Variation Ratios ($VR$) & 82.76 & \textbf{90.29} & 94.06 & 41.08 & \textbf{56.97} & 69.45 \\
            \textbf{Ours} ($H_{ens}$) & 82.88 & 90.15 & \textbf{94.33} & 40.87 & 56.94 & \textbf{70.12} \\
            \hline
        \end{tabular}
	\end{center}
	\vspace{-0.5cm}
\end{table*}

\begin{table*}[!t]
	\begin{center}
    \caption{\textbf{Comparison to Ensembles and single models on CIFAR and ImageNet:} Validation accuracy (in \%) of DPEs compared to deterministic models and ensembles using $L_2$ regularization. $R$ refers to random and $H_{ens}$ is the predictive entropy as defined in Eq. \ref{e:h_ens}. We randomly sample the initial 2k for CIFAR and the initial 12.8k for ImageNet. For CIFAR datasets, DPEs consistently improve active learning performance indicating that DPEs produce better uncertainty estimates. For ImageNet, we see similar performance of DPEs compared to $L_2$ Ensembles, giving a 3\%-4\% improvement in performance over the random baselines by employing active learning.}
    \label{t:active_32}
    \vspace{-0.25cm}
		\begin{tabular}{c|c|c|c|c||c|c|c||c|c}
			\hline
			\multicolumn{2}{c|}{\small \textbf{Dataset}}&\multicolumn{3}{c||}{ \textbf{\small{CIFAR-10}}}&\multicolumn{3}{c||}{\small \textbf{CIFAR-100}}&\multicolumn{2}{c}{\small \textbf{ImageNet}}\\
			\hline
			\small \textbf{Model} & \small \textbf{Acq.} & \small \textbf{4k (8\%)} & \small \textbf{8k (16\%)} & \small \textbf{16k (32\%)}& \small \textbf{4k (8\%)} & \small \textbf{8k (16\%)} & \small \textbf{16k (32\%)}&\small\textbf{25.6k (2\%)} & \small \textbf{51.2k (4\%)}\\
			\hline
			\small Deterministic & $R$ & 80.43 & 86.40 & 90.41 & 32.42 & 48.80 & 63.87 & 38.02 & 56.22 \\
			& $H_{ens}$ & 81.40 & 89.16 & 93.57 & 31.77 & 50.72 & 65.18& 37.61 & 55.81\\\cline{1-10}
			\small $L_2$ Ensemble & $R$ & 80.72 & 86.77 & 91.12 & 39.82 & 55.13 & 66.43 & 49.01 & 64.15 \\
            & $H_{ens}$& 82.41 & 90.05 & 94.13 & 40.49 & 56.89 & 69.68 & \textbf{52.95} & 67.25\\\cline{1-10}
            \small \textbf{Ours} (DPE) & $R$ & 80.60 & 86.80 & 91.08 & 39.57 & 54.92 & 66.65& 48.97 & 64.06 \\
            & $H_{ens}$ & \textbf{82.88} & \textbf{90.15} & \textbf{94.33} & \textbf{40.87} & \textbf{56.94} & \textbf{70.12} & 52.89 & \textbf{67.28} \\
            \hline
    
    \end{tabular}
	\end{center}	
	\vspace{-0.5cm}
\end{table*}
\subsection{Results}

\noindent
\textbf{Growth Parameter.} In our first experiment, we focus on analyzing variations of the growth parameter $b$. This parameter directly affects the number of times the model must be retrained, and therefore has large implications on the time needed to reach the final annotation budget $B$ and complete the experiment. For this experiment, we consider the CIFAR-10 dataset and an overall budget of $B=16k$ samples. As a baseline, we use random sampling as an acquisition function with $b=2k$, training the DPE 8 times. For reference, we also compute the upper bound achieved by training the DPE with all 50k samples in the dataset. We consider three variations of the growth parameter $b$ with the $H_{ens}$ acquisition function: (i) \textbf{Linear-8}: we set $b=2k$, training the DPE 8 times; (ii) \textbf{Linear-4}: we set $b=4k$, training the DPE 4 times; and (iii) \textbf{Exponential-4}: we initially set $b=2k$, and then iteratively double its value ($b=2k, b=4k$ and $b=8k$ for the 3 remaining active learning iterations). Fig. \ref{f:c10} shows the error bars over three trials. Linear-8, which is computationally the most demanding, only marginally outperforms the other approaches. This is probably due to the combination of two factors: (i) less correlation among the samples added (due to the lower growth parameter), and (ii) less reliance on models trained with very small amounts of data to select a large number of samples. It is able to achieve 99.2\% of the accuracy of the upper bound with just 32\% of the labelling effort.

When the model is only retrained 4 times, both Linear-4 and Exponential-4 data addition methods achieve similar performance levels. From a computational point of view, the exponential approach is more efficient as the dataset size for the first three training iterations is smaller (2k, 4k and 8k compared to 4k, 8k and 12k).


\noindent
\textbf{Acquisition Function.} In a second experiment, we analyze the relevance of the acquisition function, a key component in the active learning pipeline. To this end, we compare $H_{ens}$ (as defined in Eq. \ref{e:h_ens}) with five other acquisition functions: random ($R$), categories first entropy ($H_{cat}$), mutual information ($MI$), {variance} ($V$) and {variation ratios} ($VR$). The formulation of these functions and the one used in our approach is listed in Table \ref{t:acquisition}.

Categories first entropy is the sum of the entropy of each of the members of the ensemble. Mutual information, also known as Jensen-Shannon divergence or information radius, is the difference between $H_{ens}$ and $H_{cat}$ \cite{shyam2018model}. This function is appropriate for active learning as it has been shown to distinguish \textit{epistemic uncertainty} (i.e., the confusion over the model's parameters that apply to a given observation) from \textit{aleatoric uncertainty} (i.e., the inherent stochasticity associated with that observation \cite{kendall2017uncertainties}). Aleatoric uncertainty cannot be reduced even upon collecting an infinite amount of data, and can lead to inefficiency in selecting samples for an uncertainty based active learning system. Variance measures the {inconsistency} between the members of the ensemble, which also aims to isolate epistemic uncertainty \cite{smith2018understanding}. Finally, variation ratios is the number of non-modal predictions (that do not agree with the mode, or majority vote) made by the ensemble, and can be thought of as a discretized version of variance \cite{chitta2018adaptive}.

Our results for this experiment are listed in Table \ref{t:CIFAR-10}. Each result corresponds to the mean of three trials. Intuitively, functions focused on epistemic uncertainty should perform better than simpler ones. However, as shown in our experiments, in practice, there is no significant difference in performance among these acquisition functions for the chosen tasks. Our choice of using a simple entropy based acquisition function ($H_{ens}$) provides competitive results to the others, and outperforms them at larger annotation budgets.

\noindent
\textbf{Comparison to $L_2$ regularization.} Finally, we focus on comparing DPEs to ensembles trained using standard $L_2$ regularization, the existing state-of-the-art for visual active learning \cite{beluch2018power}, as well as deterministic models using $L_2$ regularization. To this end, we run experiments on CIFAR-10, CIFAR-100 and ImageNet. For CIFAR, we use an initial growth parameter of 2k samples, and compare performances at 4k, 8k and 16k samples (Exponential-4) with the $H_{ens}$ acquisition function. For ImageNet, we use a slightly different experimental setting to account for the different scale of the dataset, comparing performances at 2\% and 4\% of the data (25.6k and 51.2k samples). A summary of these results along with baselines using random data sampling ($R$) is listed in Table \ref{t:active_32}. Each result corresponds to the mean of three trials. As shown, active learning with DPEs clearly outperforms random baselines, and provides better results compared to active learning methods with single models and ensembles with standard $L_2$ regularization. We ran significance analysis using Z-tests over these results, which show significant differences for the experiments on CIFAR-10 with all 3 annotation budgets, and CIFAR-100 for the largest annotation budget, with our approach outperforming all others. This indicates that the benefits of our approach scale with the annotation budget, which is a particularly appealing property for large-scale labeling projects.

\begin{figure}[!t]
\centering
\includegraphics[width=1\columnwidth]{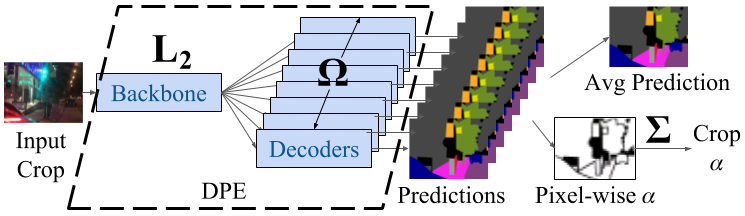}
\caption{Setup for active semantic segmentation, with a single encoder and multiple decoder heads. Uncertainty in the crops is measured by summing the acquisition functions over all pixels.}
\label{f:seg_arch}
\end{figure}

\begin{figure}[!t]
\centering
\includegraphics[width=1\columnwidth]{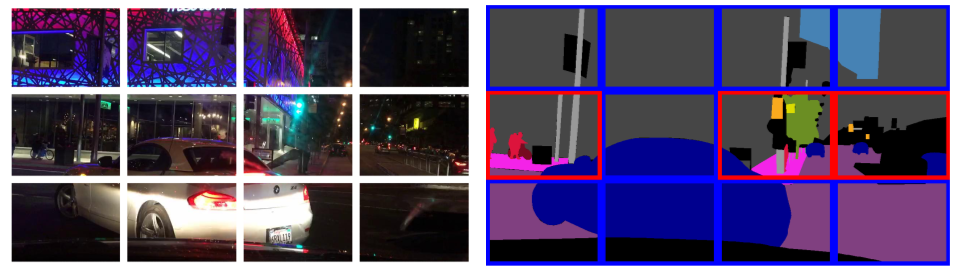}
\caption{Example of crops queried on a $4 \times 3$ grid for semantic segmentation. Some parts of the image (outlined in \textcolor{red}{red}) can be more interesting than others (outlined in {\color{blue}blue}). Querying labels for crops leads to a more effective utilization of the annotation budget than querying for images (best viewed in color).}
\label{f:split}
\end{figure}

\section{Active Semantic Segmentation}
The goal of semantic segmentation is to assign a class label to every pixel in an image. Generating high-quality ground truth at the pixel level is a time-consuming and expensive task. For instance, labeling an image was reported to take 60 minutes on average for the CamVid dataset \cite{brostow2008segmentation}, and approximately 90 minutes for the Cityscapes dataset \cite{cordts2016cityscapes}. Due to the substantial manual effort involved in producing high-quality annotations, segmentation datasets with precise and comprehensive label maps are \textit{orders of magnitude} smaller than classification datasets, despite the fact that collecting unlabeled data is not very expensive. In this section, we introduce our framework for applying DPEs for active semantic segmentation, which can help make better use of an annotation budget when building segmentation datasets. 

The proposed framework is outlined in Fig. \ref{f:seg_arch}. The DPE consists of a convolutional backbone as an encoder, and an ensemble of decoders. Only a small subset of the parameters and activations of the network are associated with these decoder layers. As a result, memory requirements for training scale linearly in only this subset of parameters. We focus on analyzing uncertainty of higher-level semantic concepts associated with deep features in the decoder, since we train the decoder with KL regularization while the encoder is trained using the common $L_2$ regularization.

\begin{figure}[!t]
\centering
\includegraphics[width=\columnwidth]{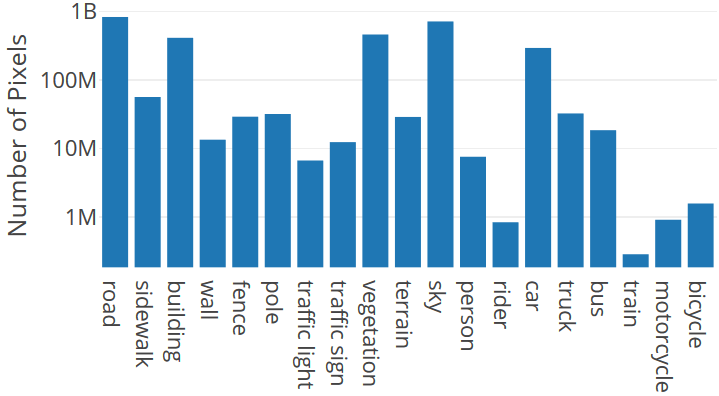}
\caption{\textbf{BDD100k:} Number of pixels of each class in the training set. While some classes (train, motorcycle, rider) have less than a million pixels, others (road, sky) have nearly a billion.}
\label{f:bdd}
\end{figure}

\begin{table}[!t]
	\begin{center}
    \caption{\textbf{BDD100k:} mIoU (in \%) comparing the proposed approach to ensembles with $L_2$ regularization for active segmentation. Initial 3.3k crops are randomly sampled. In our setup, DPEs improve upon the mIoU of ensembles by up to $1\%$. Acq. stands for acquisition function where $R$ refers to random and $H_{ens}$ is defined in Eq. \ref{e:h_ens}.}
    \label{t:active_seg}
		\begin{tabular}{c|c|c|c|c}
			\hline
			\small \textbf{Model} & \small \textbf{Acq.} & \footnotesize \textbf{6.7k (8\%)} & \footnotesize \textbf{13.4k (16\%)} & \footnotesize \textbf{26.9k (32\%)}\\
            \hline
			 \small $L_2$ Ensemble & $R$ & 45.25 & 47.00 & 48.43 \\
             & $H_{ens}$ & 45.60 & 47.62 & 49.14 \\
             \hline
            \small \textbf{Ours} (DPE) & $R$ & 45.37 & 46.92 & 48.41 \\
             & $H_{ens}$ & \textbf{45.80} & \textbf{48.10} & \textbf{50.12} \\
            \hline
        \end{tabular}
	\end{center}
	\vspace{-0.5cm}
\end{table}

In our experiments, as shown in Fig. \ref{f:split}, we split a training image into a $4 \times 3$ grid of cropped sections. To measure the uncertainty of a crop, we sum the acquisition function $\alpha$ over each pixel in the crop. We then make queries for crops rather than querying for entire images, which could lead to more efficient utilization of the annotation budget. Since the loss used to train semantic segmentation models is computed independently at each pixel, training these models from partial annotations of crops is straightforward. 
\begin{table*}[!t]
  \begin{center}
  \small
  \caption{\textbf{BDD100k:} Active learning results after training with 26.9k image crops (32\% of the dataset), showing IoU for each class and mean IoU. For comparison, we include the upper bound results obtained using a model trained with all 84k training crops. For IoU differences of at least 1\%, we mark the better result in \textbf{bold}. We observe improvements in performance on foreground classes (person through bicycle), especially those with very few instances in the training distribution (like motorcycle).}
  \label{t:bdd}
  \setlength{\tabcolsep}{1.35pt}
  \begin{tabular}{c|cccccccccccccccccccgg}
    \hline
    \textbf{Data Sampling}&\rot{road} & \rot{sidewalk} & \rot{building} & \rot{wall} & \rot{fence} & \rot{pole} & \rot{traffic light} & \rot{traffic sign} & \rot{vegetation} & \rot{terrain} & \rot{sky} & \rot{person} & \rot{rider} & \rot{car} & \rot{truck} & \rot{bus} & \rot{train} & \rot{motorcycle} & \rot{bicycle} & \rot{\textbf{mIoU @ 26.9k}} & \rot{\textbf{mIoU @ 84k}} \\ 
    \hline
	Upper Bound (DPE + All 84k) &	92.1&    56.3&    81.6&    20.7&   40.3&42.5&    44.9&    43.6&    84.3&    45.0&    93.7&56.1&    25.1&    86.6&    38.9&    56.9&    0.0& 38.6&    42.1&  - &  52.1    \\
    \hline
    \hline
	Random (DPE + $R$) &    91.2&    54.7&    80.2&    17.3&  \textbf{38.0}&    38.4&    40.8&    40.3&    83.6&   43.3&    92.9&    53.6&    18.5&    84.8&    33.3&   47.2&    0.0&    16.9&    44.9&    48.4 & - \\
		\textbf{Ours} (DPE + $H_{ens}$) &	91.3&    55.0&    80.3&   {18.4}&    36.4&    38.2&    40.0&    39.9&   83.5&    42.6&    93.0&    \textbf{55.1}&   \textbf{20.1}&    85.5&    34.1&    \textbf{52.2}&   0.0&    \textbf{36.8}&    {50.0}&   \textbf{50.1} & - \\
		{Target Signs} (DPE + $V_w$) &	91.1&55.2&80.1&\textbf{22.0}&36.9&38.7&40.5&\textbf{41.9}&83.6&41.3&93.0&53.2&16.9&85.3&31.8&48.0&0.0&34.1&49.8&49.7&-\\
    \hline
  \end{tabular}
  \end{center}
  \vspace{-0.65cm}
\end{table*}
Additionally, we consider a special use case where the performance in specific classes is critical. To that end, we define an acquisition function suitable for targeting specific classes based on a class weighting vector, $\textbf{w}$. This class-weighted acquisition function is defined as
\begin{equation}
V_w = \sum_{k \in K} \textbf{w}_k \underset{e \in E}{\text{Var}}(\textbf{p}_k^{(e)}).
\label{e:v_m}
\end{equation}

\noindent
\textbf{Experiments.} We set up the active segmentation experiments in a similar way to the Exponential-4 scheme we used for classification. We use iterations at 8\%, 16\% and 32\% of the data, retraining from scratch in every new iteration.

\noindent
\textbf{Dataset.} We experiment with the BDD100k \cite{yu2018BDD} dataset, which consists of 7000 training images and 1000 validation images of resolutions $1280 \times 720$ provided with fine pixel annotation over $K=19$ classes. Since we have 12 crops per image (see Fig. \ref{f:split}), we perform active learning over the set of 84k crops in the training partition. There is significant class imbalance in this dataset, as shown in Fig. \ref{f:bdd}.

\noindent
\textbf{Network Architecture.} We use a fully convolutional encoder-decoder network in our experiments \cite{long2015fully}. For the encoder, we dilate the last 2 blocks of a ResNet-18 backbone so that the encoded features are 1/8 of the input resolution \cite{yu2016multiscale}. For each decoder, we use a $1 \times 1$ convolutional layer with 64 kernels followed by another $1 \times 1$ convolution to a $K=19$ dimensional vector before applying the softmax activation.

\noindent
\textbf{Implementation Details.} We use a batch size of 8, learning rate of 0.0001, $\beta$ of $10^{-4}$, and patience of 10 epochs. Our network is initialized by pre-training on the same 19-class segmentation task with the Cityscapes dataset \cite{cordts2016cityscapes}. We resize the Cityscapes images to a resolution of $1536 \times 768$, and train on random crops of size $320 \times 320$ for 100 epochs, with a batch size of 16 and an exponentially decaying learning rate initialized at 0.001.

To maintain simplicity in our setup, we avoid data augmentation, class-weighted sampling, class-weighted losses, or auxiliary losses at intermediate layers. We check the performance of this setup by training a fully-supervised version of our model on the entire dataset, which achieves a mean IoU of 52.1\%, just 1\% less than the baseline result provided for a model of similar architecture but slightly more parameters with the BDD100k dataset \cite{yu2018BDD}. 

\subsection{Results}

The mean segmentation IoUs over 3 trials are shown in Table \ref{t:active_seg}. As with classification, we observe clear overall benefits of DPEs in the semantic segmentation setup, with nearly a 2\% improvement in mean IoU over the random baseline and 1\% over ensembles at 26.9k training crops. Again, the relative gap with competing methods is initially small, but grows steadily as the size of the labeled set increases. {DPEs recover 96.2\% of the performance of the upper bound with just 32\% of the labeling effort.} 

We compare our class-wise performance at 26.9k crops to the random baseline and fully-supervised upper bound in Table \ref{t:bdd}. We observe the largest margins of improvement with our method for foreground classes with fewer instances in the training data. On the motorcycle class, of which there are only 240 training instances, we see a difference in IoU of nearly 20\%.

The last row in Table \ref{t:bdd} shows our results using $V_m$, as defined in Eq. \ref{e:v_m}, for targeting the traffic sign class. Specifically, we set $\textbf{w}_k$ for traffic sign to 1 and the remaining weights to 0. As shown, this results in a 2\% improvement in IoU over the $H_{ens}$ acquisition function for the traffic sign class, without affecting significantly the IoU improvements on other classes (even improving by 3.6\% on the wall class). We note that the gap to the upper bound for traffic signs is improved to 1.7\% from 3.7\%, which is a 54.1\% relative performance gain. 

\section{Conclusion}
In this paper, we introduced Deep Probabilistic Ensembles (DPEs) for uncertainty estimation in deep neural networks. The key idea is to train ensembles with a novel KL regularization term as a means to approximate variational inference for BNNs. Our results demonstrate that DPEs improve performance on active learning tasks over baselines and state-of-the-art active learning techniques on three different image classification datasets. Importantly, contrary to traditional Bayesian methods, our approach is simple to integrate in existing frameworks and scales to large models and datasets without impacting performance. Moreover, when applied to active semantic segmentation, DPEs yield up to 20\% IoU improvement in underrepresented classes.

{\small
\bibliographystyle{ieee}
\bibliography{egbib}
}

\end{document}